\documentclass{article}

\PassOptionsToPackage{table}{xcolor}
\usepackage{microtype}
\usepackage{graphicx}
\usepackage{subcaption}
\usepackage{booktabs} % for professional tables
\usepackage[breakable]{tcolorbox}
\usepackage{dsfont}
\usepackage{tabularx}
\usepackage{pifont}
\usepackage{caption}
\usepackage{adjustbox}
\usepackage{hyperref}
\newcolumntype{Y}{>{\centering\arraybackslash}X}
\usepackage[accepted]{icml2025}

\usepackage{amsmath}
\usepackage{amssymb}
\usepackage{mathtools}
\usepackage{amsthm}

\usepackage[capitalize,noabbrev]{cleveref}
\usepackage{array} % Add this to your preamble

\icmltitlerunning{Tell me why: Visual foundation models as self-explainable classifiers}

\begin{document}

\twocolumn[
\icmltitle{Tell me why: \\Visual foundation models as self-explainable classifiers}



\icmlsetsymbol{equal}{*}

\begin{icmlauthorlist}
\icmlauthor{Hugues Turb\'e}{unige,hug}
\icmlauthor{Mina Bjelogrlic}{unige,hug}
\icmlauthor{Gianmarco Mengaldo}{singapore}
\icmlauthor{Christian Lovis}{unige,hug}
\end{icmlauthorlist}

\icmlaffiliation{unige}{Department of Radiology and Medical Informatics, University of Geneva, Geneva, Switzerland}
\icmlaffiliation{hug}{Division of Medical Information Sciences, Geneva University Hospitals, Geneva, Switzerland}
\icmlaffiliation{singapore}{Department of Mechanical Engineering, College of Design and Engineering, \\National University of Singapore, Singapore}

\icmlcorrespondingauthor{Hugues Turb\'e}{hugues.turbe@unige.ch}

\icmlkeywords{Machine Learning, Intepretability, Explainability, Visual Classification, VFM, Visual Foundation Model}

\vskip 0.3in
]



\printAffiliationsAndNotice{}  % leave blank if no need to mention equal contribution

\begin{abstract}
Visual foundation models (VFMs) have become increasingly popular due to their state-of-the-art performance. 
However, interpretability remains crucial for critical applications. 
In this sense, self-explainable models (SEM) aim to provide interpretable classifiers that decompose predictions into a weighted sum of interpretable concepts. 
Despite their promise, recent studies have shown that these explanations often lack faithfulness. 
In this work, we combine VFMs with a novel prototypical architecture and specialized training objectives. 
By training only a lightweight head (approximately 1M parameters) on top of frozen VFMs, our approach (ProtoFM) offers an efficient and interpretable solution. 
Evaluations demonstrate that our approach achieves competitive classification performance while outperforming existing models across a range of interpretability metrics derived from the literature. Code is available at \url{https://github.com/hturbe/proto-fm}.

\end{abstract}

\section{Introduction}
Deep neural networks have shown impressive results in vision tasks, including segmentation and classification. 
Recent progress has been largely driven by the development of visual foundation models (VFMs), that is, models trained on vast image datasets that serve various downstream tasks. 
As VFMs are increasingly applied in diverse domains, interpretability is becoming crucial for critical fields like medicine and remote sensing. 
Given the ambiguity around what interpretability means~\cite{lipton2018mythos}, in this work we focus on the ability to reflect the influence of the input features on the model's prediction. 
Besides legal requirements \cite{metikovs2024right}, the absence of interpretability has often been pointed out as a barrier to the adoption of deep learning models in various domains, such as the healthcare sector~\cite{reddy2022explainability}. 
Beyond the adoption challenges, interpretability may allow for scientific discovery where a model might have discovered some ``rules"  in the data unknown to the scientific community that can be uncovered through explaining the model's predictions~\cite{mengaldo2025explainableaiscientificmethod}.

Methods for model interpretability can be broadly categorized into two paradigms based on the stage of model development where interpretability is incorporated: (i) the intrinsic paradigm, and (ii) the post-hoc paradigm \cite{madsen2024interpretability}. 
Post-hoc methods aim to provide explanations for already trained models. 
These methods are often model-agnostic (e.g., SHAP~\cite{NIPS2017_8a20a862}), and do not interfere with model training or performance. 
However, these methods are often criticized for their lack of faithfulness~\cite{NEURIPS2018_294a8ed2,NEURIPS2019_fe4b8556,turbe2023evaluation,wei2024revisiting}, raising significant concerns about their reliability in critical applications. This has led to growing advocacy for the intrinsic paradigm~\cite{rudin2019stop}, which involves models designed to be interpretable by design, which are also referred to as self-explainable models (SEM).

A notable direction within SEM is the development of part-prototype models, which aim to decompose predictions into a weighted sum of interpretable concepts. 
Although these models are theoretically designed to provide consistent explanations, recent findings indicate that the explanations they generate often lack faithfulness. 
Specifically, these models tend to mislocalize critical regions of the input that are important for classification~\cite{sacha2024interpretability,Carmichael_2024_WACV} and fail to represent coherent concepts in the input space~\cite{hoffmann2021looks}, aspects that undermine their interpretability. 
Another general criticism of the intrinsic paradigm is that previous approaches are not designed to leverage pre-trained models, such as vision foundation models (VFMs)~\cite{madsen2024interpretability}.

In this work, we aim to address the challenges just described that affect SEMs . 
We achieve this goal by designing a new architecture, ProtoFM, depicted in Figure~\ref{fig:model}, that (i) significantly improves the quality of the explanations provided, and (ii) allows using frozen VFMs.  

One of the key motivations behind using existing VFMs, such as Dinov2~\cite{oquabdinov2}, AM-RADIO~\cite{ranzinger2024radio}, and SAM~\cite{kirillov2023segment}, is their state-of-the-art performance on a variety of downstream tasks, including classification and segmentation in general image domains~\cite{oquabdinov2,ranzinger2024radio}, and specialized tasks -- e.g., RAD-DINO~\cite{perez2025exploring} for chest radiography, and SkySense~\cite{guo2024skysense} for remote sensing.

We argue that using frozen VFMs enables efficient training on diverse classification tasks with a minimal number of parameters trained. Furthermore, VFMs trained with patch-level objectives, yield strong local representations, as evidenced by their high performance in segmentation~\cite{oquabdinov2,perez2025exploring}. This property is important to addresses the critical issue of \emph{spatial alignment} in prototypical models.

% Significance Statement Box
\begin{tcolorbox}[colback=blue!5!white, colframe=blue!75!black, title=Contributions, breakable]
\begin{enumerate}
\item Introduce a novel architecture which allows to leverage frozen VFMs providing a lightweight approach ($\approx 1$M trained parameter). 
The evaluation demonstrates that the architecture is competitive in terms of classification performance  achieves SOTA performance across the range of interpretability desiderata derived from the literature.
\item We extensively evaluate a number of prototypical models from the literature, highlighting important issues regarding the correctness and contrastivity of the explanations obtained with these models, justifying the introduction of the novel architecture.
\end{enumerate}
\end{tcolorbox}
\vspace{-5pt}

\section{Related work}
The development of Prototypical Part Network research was initiated by the ProtoPNet architecture ~\cite{protopnet}. ProtoPNet works by comparing parts of an image to learned prototypes, which are meant to represent semantically coherent concepts. This approach allows the model to make predictions based on the similarity of image parts to these prototypes, offering a form of reasoning that is qualitatively similar to human decision-making.

Several challenges with the original ProtoPNet have led to the development of more advanced models. The original ProtoPNet used a fixed number of class-specific prototypes. This means that each class had a pre-determined number of prototypes associated with it, which could lead to large explanations and redundant prototypes~\cite{nauta2021looks}. ProtoPShare was developed to address this issue by enabling prototypes to be shared across different classes, reducing the total number of prototypes needed and allowing the model to find similarities between classes~\cite{rymarczyk2021protopshare}. ProtoPool built on this work and introduced a fully differentiable assignment of prototypes to classes, allowing end-to-end training of the model~\cite{rymarczyk2022interpretable}. 

Other approaches have explored different prototype representations. Deformable ProtoPNet introduced spatially flexible prototypes~\cite{donnelly2022deformable}. ST-ProtoPNet aimed to improve classification accuracy by learning support and trivial prototypes, drawing an analogy with Support Vector Machine theory~\cite{Wang_2023_ICCV}. On a different note, ProtoTree looked at replacing the final linear layer with a decision tree~\cite{nauta2021neural}, while PIP-Net focused on producing sparse explanations~\cite{nauta2023pip}. Most of the presented works described above use different variations of CNN backbones, but the ProtoVit architecture recently looked at leveraging Vision Transformers (ViT)~\cite{mainterpretable}.

\begin{table*}[t]
\centering
\caption{\textbf{Properties and associated metrics for evaluating interpretability}. $^*$ denotes metrics from the FunnyBirds framework. $^\dagger$ indicates metrics from the literature adapted to the FunnyBirds dataset}
\label{tab:interpretability_metrics}
\rowcolors{2}{gray!15}{white} % Alternating row colors
\begin{tabular}{m{0.15\linewidth}m{0.55\linewidth}m{0.25\linewidth}}
\hline
\textbf{Property} & \textbf{Definition} & \textbf{Associated Metrics} \\ \midrule
\textbf{Correctness} &  The explanation faithfully represents the model’s behavior. & SD$^*$ \\ 
\textbf{Completeness} & The model’s behavior is fully captured by the explanation. & CSDC$^*$, PC$^*$, DC$^*$, D$^*$ \\ 
\textbf{Contrastivity} &  Discriminative parts are correctly captured by the explanations & TS$^*$ \\ 
\textbf{Consistency} &  Prototypes are consistent in the input space. & Consistency$^\dagger$ \\ 
\textbf{Stability} & Prototype attribution should be stable under small perturbations. & Stability$^\dagger$ \\
\textbf{Compactness} &  The explanation is compact to be intelligible by the user. & Global$^\dagger$ and Local size$^\dagger$ \\
\textbf{Composition} & The explanation presentation should reflect the model’s behavior. &  SEC  \\ \bottomrule
\end{tabular}
\end{table*}

While part-prototype networks are theoretically designed to be interpretable, researchers have investigated their interpretability and identified several limitations that can undermine this promise. Studies have shown that such models may exhibit: (i) a \textit{semantic gap}, where prototypes fail to consistently represent the same concepts across different images~\cite{hoffmann2021looks,kim2022hive,nauta2023pip}, and (ii) \textit{spatial misalignment}, where the pixels used by the model for predictions are not correctly localized~\cite{carmichael2024pixel,sacha2024interpretability}.

While several studies have examined specific aspects of interpretability evaluation~\cite{gautam2022protovae,carmichael2024pixel,huang2023evaluation}, none have provided a holistic and quantitative assessment of prototypical models' interpretability that addresses its multifaceted nature. The Co-12 properties were recently introduced as a comprehensive framework for evaluating explanation quality~\cite{nauta2023anecdotal}. These properties have been designed for interpretable methods in general, including both post-hoc interpretability methods and SEMs. We summarize the most important properties in~\cref{tab:interpretability_metrics}. They encompass many desiderata that have been independently formulated in the literature focused on SEMs with metrics such as \textit{Prototypical part Location Change (PLC)}~\cite{sacha2024interpretability} or \textit{Relevance Ordering Test (ROT)}~\cite{carmichael2024pixel} which aim to evaluate the spatial alignment and fall under the \textit{Correctness property}. Metrics to evaluate the consistency, stability and compactness desiderata have also been proposed in the literature~\cite{huang2023evaluation,nauta2023pip}.   

However, the metrics and their evaluation introduced to evaluate Prototypical models suffer from two main issues: i) Distribution shift under pixel ablation and ii) Lack of precise part annotations. Regarding the first issue, interpretability evaluations are typically conducted by removing pixels that are considered important by an interpretability method and assessing the resulting change in the model's prediction. However, this ablation creates a distribution shift between the training dataset and the dataset used for the interpretability evaluation which might alter the evaluation~\cite{hooker2019benchmark}. The second aspect is related more specifically to the evaluation of SEM which often aims to measure the consistency of the prototype used for the classification based on part annotations (e.g. beak, wing in the CUB dataset)~\cite{nauta2023pip,carmichael2024pixel,huang2023evaluation}. However, the datasets used for these evaluations do not include precise annotations for individual parts. As a result, the evaluation relies on estimating part localization by placing a fixed-size box around a point that represents the part's position.Evaluation performed with these annotations considers only the top patches activated by a prototype which is then compared to the fixed-size box. This approach has been criticized as not correctly reflecting the model's decision process~\cite{Carmichael_2024_WACV}. 

The FunnyBirds dataset has been developed to evaluate a three of the Co-12 properties, namely correctness, completeness and contrastivity. We list the associated metrics from FunnyBirds in~\cref{tab:interpretability_metrics} and refer the reader to the paper introducing these metrics for more details~\cite{hesse2023funnybirds}. The developed metrics and dataset address the issues stated above and unify the evaluation of post-hoc interpretability methods and SEM models under a single framework. As part of their benchmark, 24 combinations of interpretability methods and models are evaluated. However, the only prototypical model evaluated is ProtoPNet. Recently, the methodology to compute the metrics presented in FunnyBirds for prototypical models was slightly modified, but again only ProtPNet was evaluated~\cite{oplatek2024revisiting}. 

Next, we present the novel architecture along with a set of metrics to improve the overall explainability of self-explainable models and thoroughly evaluate the explanations provided by these models.

\section{Methodology}

\textbf{Problem setting} We consider the classification task that consists of mapping an image $X\in H\times W \times C$ to a labelled target $\mathcal{Y} \in \mathbb{N}^D$ where $H$, $W$, $C$ represent, respectively, the height, width, and number of channels of the input image, and $D$ is the number of classes.

\subsection{Model architecture}

\begin{figure*}
    \centering
    \includegraphics[width=0.95\linewidth]{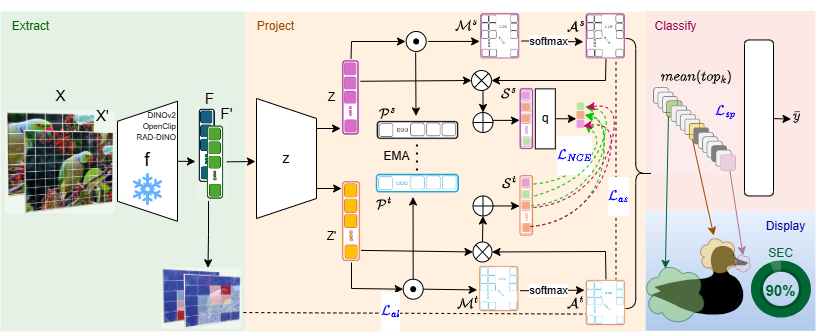}
    \caption{Model architecture. The model is composed of a frozen VFM followed by a projector and classification head  in order to classify images from a set of learned concepts.}
    \label{fig:model}
\end{figure*}

As depicted in~\cref{fig:model}, our model ProtoFM leverages a frozen VFM as a visual feature extractor $f$ mapping image $x$ to patch embedding $F_i \in \mathbb{R}^C $ for patch index $i \in [1,\dots, I]$, and $I= \frac{H}{s} \cdot \frac{W}{s}$ with $s$ indicating the patch size of the VFM and $c_f$ the embedding dimension. Similarly a version augmented with geometric and color transformation $x'$, is mapped to $F'$. 

A projector $z$ then maps the feature from the backbone into an embedding space, of dimension $c_z$, such that the image and its augmented version are mapped to $Z=z(F)$ and $Z'=z(F') \in \mathbb{R}^{c_z \times I}$. A cosine similarity is then computed between $Z$ and a set of trainable prototypes $\mathcal{P}^{\{s,t\}} = \left \{p_n \in \mathbb{R}^{c_z}\right \}$, with $n \in[1,\cdots,N]$ where $N$ is the number of prototypes. The prototypes aim to represent the concepts that can decompose the image and be used to classify the latter. The $s$ and $t$ exponents respectively refer to the student and teacher prototypes, with the latter being updated through an exponential moving average (EMA) of the student prototype. 
 
To improve the consistency of the prototypes assignment, we follow a student-teacher approach similar to the segmentation model proposed in \textit{SmooSeg}~\cite{smooseg}. Two masks, $\mathcal{M}^s$ and $\mathcal{M}^t$ attributing pixels from the image to a prototype are computed by measuring the cosine similarity between the prototypes and the projections as follows: \begin{equation}
    \mathcal{M}^s=\cos(\text{sg}(Z);\Phi_s) \quad; \quad \mathcal{M}^t=\cos(Z';\text{sg}\left(\Phi_t)\right)
\end{equation}
where sg stands for the stop-gradient operations. To compare the two masks which come from two different views of a given image, the two masks are aligned into overlapping regions using  RoIAlign~\cite{he2017mask} to obtain $\mathcal{\tilde{M}}^s$ and $\mathcal{\tilde{M}}^t \in \mathbb{R}^{I \times N} $. The prototypes soft-assignments are then computed for the aligned and non-aligned mask: \begin{equation}
    \mathcal{A}^{\{s,t\}} = \sigma\left( M^{\{s,t\}} /\tau\right) ; \tilde{\mathcal{A}}^{\{s,t\}} = \sigma \left( \tilde{M}^{\{s,t\}} /\tau\right)
\end{equation}
where $\sigma$ denotes the softmax operation across dimension $N$ and $\tau$ the temperature .
The prototype assignments are aggregated at the image level using a top-\( k \) mean operation, which captures the presence of each prototype within a given image by averaging its \( k \) largest activation values: 
\begin{equation}
    h^s_n = \text{mean} \big( \text{top}_k(\mathcal{M}^s_n) \big) \quad ; \quad h^t_n = \text{mean} \big( \text{top}_k (\mathcal{M}^t_n) \big)
\end{equation}
where \( h_n \) quantifies the presence of prototype \( n \) in the image. The final classification head consists of a linear classifier with weights \( W \) constrained to be positive, a design choice aimed at enhancing the model's interpretability. This linear layer processes the vector \( H = \{ h_n \in \mathbb{R} \mid n \in [1, \ldots, N] \} \) to produce an importance matrix $\mathbf{R}=\left(r_{d,n}\right) \in \mathbb{R}^{D\times N}$ with the class-specific importance score ${r}_{d,n}$ of prototype $n$ toward class $d$:
\begin{equation}
    r_{d,n} = W_{d,n} \times h_n.
\end{equation}
The final prediction for class \( d \) is obtained by summing the contributions of all prototypes relevant to that class:
\begin{equation}
    \bar{y}_d = \sum_n r_{d,n}.
\end{equation}

To improve the consistency of the prototypes representation, we further compute a prototype representation per view, taking inspiration from~\cite{wen2022self} as follows: \begin{equation}
    \mathcal{S}^s = \frac{1}{\sum_{I} \mathcal{A}_i} \sum_{I} \mathcal{A}^t_i \odot Z_i \quad \text{,} \quad  \mathcal{S}^s \in \mathbb{R}^{N \times c_z}
\end{equation}
where $\odot$ denotes the Hadamard product. This operation is repeated with the teacher branch to obtain $\mathcal{S}^t$. These representations are used to contrast the prototype representations (or slots) in a contrastive loss. The representation from the teacher branch is projected using a two-layer MLP $q$ to get $Q^t=q(S^t) \in \mathbb{R}^{N \times C_z} $. We next describe the overall optimization objective. 

\subsection{Optimization objective}
The optimization objective is composed of losses promoting a consistent and local assignment of the patches along a joint-classification objective. The different losses used in this sense are described next.

\textbf{Assignment loss:} To promote a consistent and confident assignment of the prototypes, we first aim to encourage the same patches from two different views, with both geometric and color transformations, to be assigned to the same prototype with high confidence through the assignment loss $\mathcal{L}_{as}$
 \begin{equation}
     \mathcal{L}_{as} = -\frac{1}{I} \sum_i^I \log \sum_d^D \tilde{A}^s_{i,d} \tilde{A}^t_{i,d} 
 \end{equation}

\textbf{Alignment loss:} Prototypes assignments are also aligned to the backbone through a correspondence distillation loss~\cite{hamilton2022unsupervised}: \begin{equation}
    \mathcal{L}_{cd}^{intra} = \left (\hat{F} -b\right)\hat{\mathcal{A}^t}
\end{equation}
where $\hat{F}$ denotes the intra-sample cosine correlation between the extracted feature $F$ and similarly for $\hat{\mathcal{A}}^t$ with the prototypes assignment $\mathcal{A}^t$. Following the method in~\cite{kim2024eagle}, the shift $b$ is adapted through the training (see~\cref{sec:details_alignment_loss}). The operation is further repeated by comparing a given sample with a random sample from the same batch to create $\mathcal{L}_{cd}^{inter}$. The final alignment loss $\mathcal{L}_{al}$ is:
 \begin{equation}
     \mathcal{L}_{al} = \mathcal{L}_{cd}^{intra} + \frac{1}{m} \sum_m \mathcal{L}_{cd}^{inter}
 \end{equation}
 where $m$ denotes the number of times a negative sample is drawn from the batch. 

 We note that the alignment of the prototypes assignment to the backbone resembles the smoothness loss presented in~\cite{smooseg}. However, we found that their loss does not prevent a model collapse under the pressure of the $\mathcal{L}_{assign}$ term (see~\cref{sec:details_alignment_loss} for more details).

\textbf{Contrastive loss:} We next leverage the prototype representations $\mathcal{S}^{\{s,t\}}$ and the projection $Q^t$ in a contrastive loss $\mathcal{L}_{NCE}$ as initially presented in~\cite{wen2022self}. This objective encourages a consistent representation of a given prototype across views of a given image as well as minimizing the similarity with different prototype representations. The contrastive loss $\mathcal{L}_{NCE}$ is defined as: \begin{multline}
\mathcal{L}_{NCE} \left( \mathcal{S}^s, \mathcal{S}^t \right) = \\
\frac{1}{N} \sum_{n=1}^{N} -\log 
\frac{ \mathds{1}^{n,s} \mathds{1}^{n,t} \exp \left( \bar{q}^{t}_n \cdot \bar{\mathbf{s}}^{s}_n / \tau \right) }
{ \sum_{n'} \mathbf{1}^{n,s} \mathbf{1}^{n',t} \exp \left( \bar{q}^{t}_{n'}  \cdot \bar{\mathbf{s}}^{s}_n / \tau \right) }.
 \end{multline}
where $\bar{\cdot}$ denotes $\ell_2$-normalization, $q_n$ and $s_n$ denotes respectively the entry of $S$ and $Q$ for prototype $n$ and $\mathds{1}^{n}$ is a binary indicator representing prototypes which are dominant in at least a single patch:
\begin{equation}
\mathds{1}^{n,s} = \exists_{I} \quad \text{such that} \quad \underset{N}{\arg\max} \left( \mathcal{A}^s \right) [I] = n
\end{equation}
and similarly for $\mathds{1}^{n,t}$ based on the teacher assignment $\mathcal{A}^t$.

\textbf{Sparsity loss:} To prune prototypes in the classification layer with low importance, we introduce a sparsity loss $\mathcal{L}_{sp}$ based on the Hoyer-Square (HS) regularizer ~\cite{yang2019deephoyer} applied on the importance matrix $\mathbf{R}$:

\begin{equation}
     \mathcal{L}_{sp} = \alpha \frac{\Vert\mathbf{R}\Vert^2_1}{\Vert\mathbf{R}\Vert_2^2} + \gamma \Vert \mathbf{R} \Vert _2.
     \label{eq:sparsity_loss}
\end{equation}

\textbf{Classification loss:} The classification loss is a simple cross-entropy loss between the model's prediction $\bar{y}$ and the label $\mathcal{Y}$ :
\begin{equation}
  \mathcal{L}_{CE} =    -\sum_q y_q \log \bar{y}_q
\end{equation}
This loss is applied both on the student and teacher predictions $\bar{y}^s$ and $\bar{y}^t$.

The final objective is: 
\begin{align} 
\begin{split}
\mathcal{L} &= \lambda_1 \mathcal{L}_{as} +\lambda_2 \mathcal{L}_{al} \\
&+ \lambda_3 \mathcal{L}_{NCE} + \lambda_4 \mathcal{L}_{sp} + \lambda_5 \mathcal{L}_{CE}
\end{split}
\end{align}
where $\lambda_{[1,\ldots, 5]}$ are hyper-parameters.

\subsection{Benchmark for evaluation of prototypical-part models}
\label{sec:interp_metrics}
The set of metrics used in our evaluation aims to provide a general evaluation of the explanations provided by the prototypical models. This evaluation can be decomposed into two main steps. We first evaluate all models with the metrics from the FunnyBirds metrics which cover three dimensions of the Co-12 properties, namely correctness, completeness, and contrastivity. The metrics in the framework rely on a part importance function $PI(\cdot )$ which aims to reflect the importance of the different parts in the image towards the model's prediction. We follow a similar approach as~\cite{oplatek2024revisiting} described in more detail in~\cref{sec:details_interp_metric_method} to adapt the $PI$ function to prototypical models. This set of metrics allows to derive a mean explainability score $mX$ which allows us to compare the different models to different interpretability methods which are not necessarily based on prototypical models.

In a second part, we focus on metrics specific to prototypical model. The consistency and stability score initially developed on the CUB dataset by~\cite{huang2023evaluation} were adapted to the FunnyBirds dataset to leverage the precise part annotations provided as part of this dataset. The consistency metrics evaluate whether the prototypes are consistently attributed to one of the five parts defined in the Funnybirds dataset, that is beak, eye, foot, tail and wing. The stability metrics focus on measuring whether the part attribution of a prototype is stable when the input is perturbed with noise. The idea is to evaluate whether prototype assignments change under perturbations which are invisible to human eyes. More details on these metrics are included in~\cref{sec:details_interp_metric_method}

To assess the compactness of the evaluation, we utilize the local and global size metrics from \cite{nauta2023pip}. The local size quantifies the total number of prototypes a model uses for a prediction, while the global size represents the number of prototypes with non-zero weight in the classification head. The composition property is often overlooked, with score sheets failing to indicate that the displayed prototypes contribute often for less than 50\% of the final prediction. To address this, we introduce the Score Explained by Composition (SEC) metric. We propose incorporating this metric into score sheets produced by prototypical models, as it measures the fraction of the total prediction explained by the prototypes presented in a given score sheet.

All the properties included in our benchmarks along the corresponding metrics are summarized in~\cref{tab:interpretability_metrics}.

\section{Experiments}
\subsection{Implementation details}
The proposed architecture leverages DINOv2 ViT-B/14 with registers~\cite{oquabdinov2,darcet2023vision} as well as the ViT-L from the OpenClip architecture~\cite{cherti2022reproducible} for the general datasets as our backbone. In addition, an experiment on a chest X-Ray dataset with RAD-DINO~\cite{perez2025exploring} was also performed to demonstrate the possibility of leveraging domain specific VFMs. Full experimental setups including the number of epochs for the different experiments are described in~\cref{sec:experimental_setup}.

\subsection{Classification performance}

\begin{table}[ht]
\caption{Performance comparison across SOTA models in terms of classification accuracies (Acc.), global (G.) and local (L.) size. $\dagger$ the most recent results  are reported~\cite{xue2022protopformer}} 
\label{tab:accuracy}
\begin{center}
\begin{small}
\begin{tabularx}{0.99 \columnwidth}{lYY|YY}
\toprule
                       & \multicolumn{2}{c}{CUB}      & \multicolumn{2}{c}{CARS}     \\ 
                       & Acc. \newline $\uparrow$ & G. / L. \newline size $\downarrow$  & Acc. \newline $\uparrow$  & G. / L.  \newline size $\downarrow$ \\ \midrule
DINOv2-B           & 89.6          &        & 88.2          &      \\ \midrule
ProtoPNet              & 79.2          & 2000/10      & 86.1          & 2000/10      \\
Proto Tree             & 82.2          & 202/         & 86.6          & 195/         \\
ProtoPShare            & 74.7          & 400/         & 86.4          & 480/         \\
ProtoPool              & 85.5          & 495/         & 88.9          & 195/         \\
PIP-Net                & 84.3          & 495/\textbf{4}        & 88.2          & 515/\textbf{4}       \\
ViT-NeT$^\dagger$      & 84.5          &              & 92.6          &              \\
PixPNet                & 81.8          & 2000/10      &               &              \\
ST-ProtoPNet           & 86.1          & 8000/40      & 92.7 & 8000/40      \\
ProtoViT               & 85.8          & 2000/10      & 92.6          & 2000/10      \\ \midrule
 \rowcolor{gray!15} ProtoFM  w/DINO & \textbf{86.3} & \textbf{74}/6         & 92.4          & \textbf{76}/7         \\ 
  \rowcolor{gray!15} ProtoFM  w/CLIP & 77.4 &    57/6    & \textbf{93.6  }        &   6/46      \\ \bottomrule
\end{tabularx}
\end{small}
\end{center}
\end{table}
We compare the proposed approach in term of classification performance to a non-explainable baseline and SOTA prototypical models. 
For the non-explainable baseline, we present results for one of the frozen backbone, i.e. DINOv2 ViT-B/14, with a linear classifier reporting results from the initial model publication \cite{oquabdinov2}. 
In addition, we consider a range of SOTA prototypical models, namely ProtoPNet~\cite{protopnet}, ProtoTree~\cite{nauta2021neural}, ProtoPShare~\cite{rymarczyk2021protopshare}, ProtoPool~\cite{rymarczyk2022interpretable}, PIP-Net~\cite{nauta2023pip}, ViT-Net~\cite{vit-net}, ST-ProtoPNet~\cite{Wang_2023_ICCV}, PixPNet~\cite{carmichael2024pixel}, ProtoViT~\cite{mainterpretable}.

\begin{figure}[ht]
    \centering
    \includegraphics[width=0.95\linewidth]{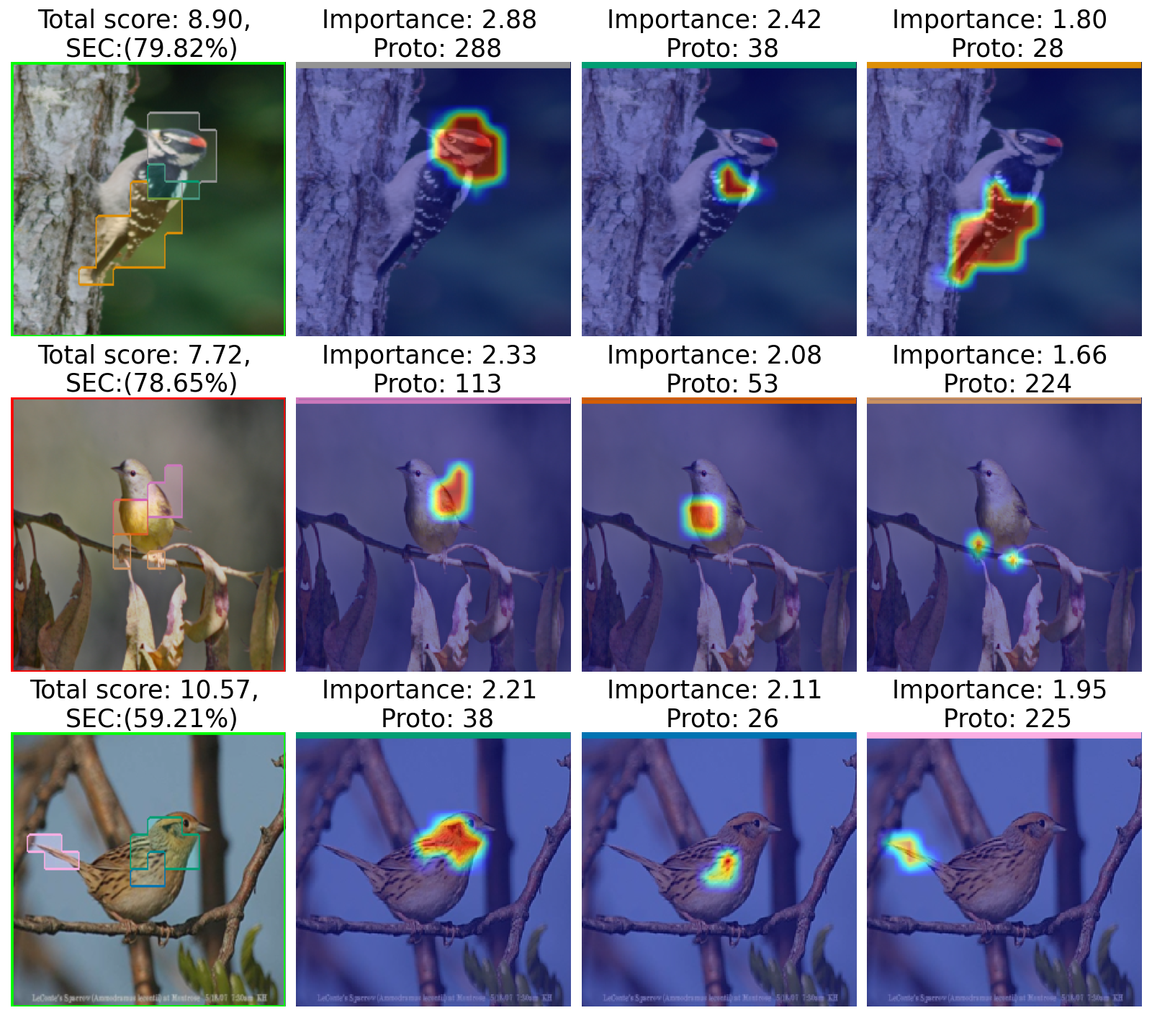}
    \caption{Score sheet for predictions on three random samples of the CUB dataset. Each row shows a
prediction on a different sample. The first column indicates the position of the top four prototypes.
Each subsequent column shows a prototype along with its importance towards the predicted class. The total score for the predicted class and the SEC metric are presented above the first column.}
    \label{fig:cub_score-sheet}
\end{figure}

\begin{figure*}[t]
    \centering
     \includegraphics[width=.9 \textwidth]{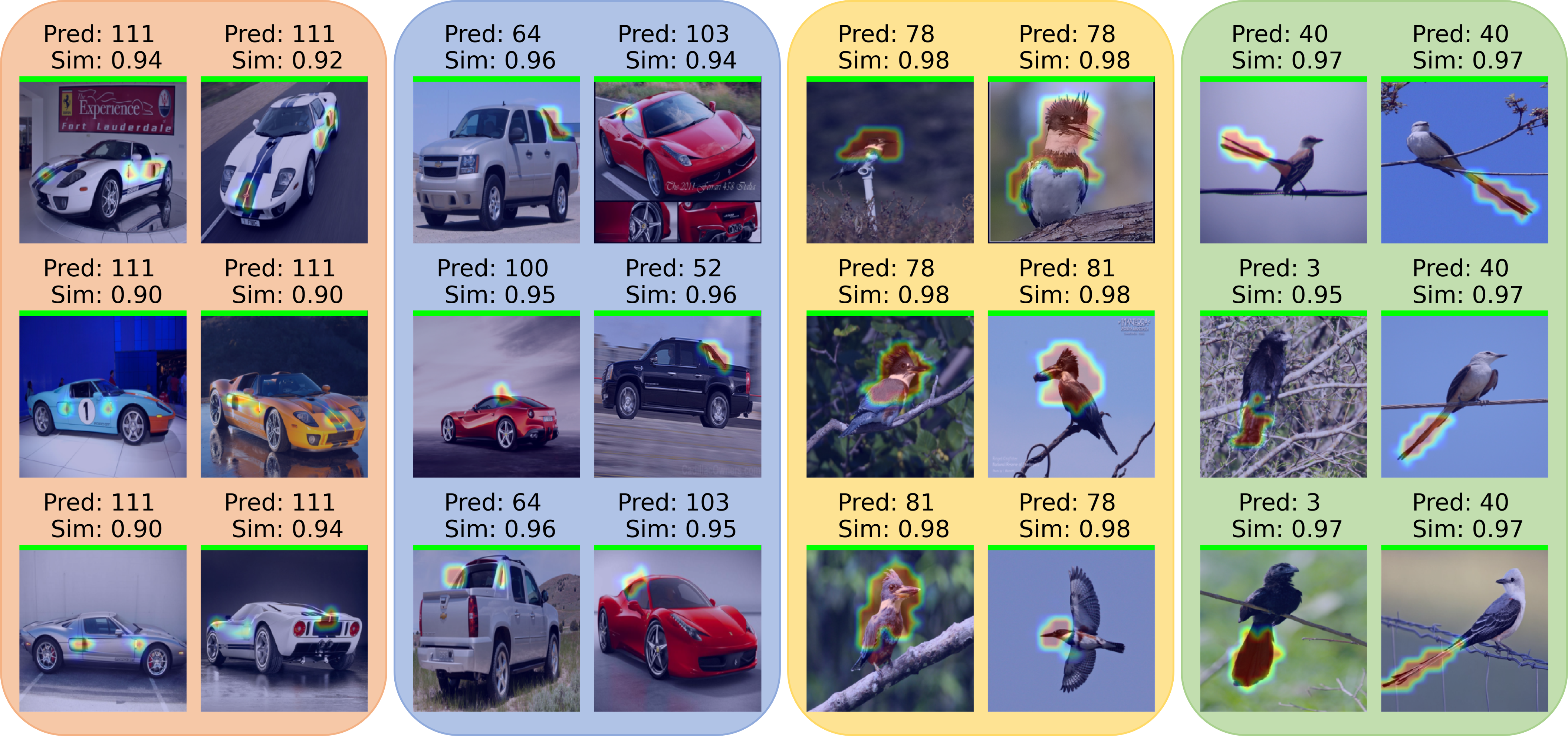}
     \caption{Nearest patches to four prototypes; two for the CARS dataset (orange and blue boxes) and two for CUB (yellow and green boxes). The predicted class along the max similarity between the prototype of interest and the patches are indicated above each image.}
     \label{fig:sim_proto}
\end{figure*}

Three datasets for image classification tasks were used to benchmark the model on general classification tasks. Two are common benchmarks: i) CUB-200-2011 \cite{wah2011caltech} (200 bird species), ii) Stanford Cars \cite{krause20133d} (196 car models). A third general dataset, Oxford-IIIT Pets (37 cat and dog species)~\cite{parkhi2012cats}, was added as it has been used to evaluate PIP-Net as well DINOv2. The proposed architecture was further evaluated on the RSNA pneumonia detection (Presence/absence of a pneumonia on chest radiographs)~\cite{shih2019augmenting} as a clinical test case.

The classification accuracy of the proposed architectures along the baselines described above is shown for CUB and CARS datasets in~\cref{tab:accuracy}. Examples of score-sheet prediction on CUB are shown in~\cref{fig:cub_score-sheet}, while a set of prototypes learned along their most similar patches are shown in~\cref{fig:sim_proto}. More results are provided in~\cref{sec:additional_visualizations} and the additional materials (~\cref{sec:additional_materials}). Results on PETS are shown in~\cref{sec:additional_results}. For baseline models, we report all results available in the literature, that is either in the paper presenting the model or in further work. 

Regarding the classification accuracy of the proposed architecture, we observe that the model achieves state-of-the-art performance on the two benchmarks, that is CUB and CARS, compared to the other prototypical models presented in the literature. Interestingly, ProtoFM w/ DINO outperforms the performance of DINOv2 with linear probing on the CARS dataset. We also note that this performance is achieved by retaining the backbone frozen and training the rest of the architecture, which is limited to only 1.3M parameters on the two datasets presented in~\cref{tab:accuracy}. Previous approaches have often leveraged pre-trained CNN backbones or even more recently ViT~\cite{mainterpretable}, but prototypical models found in the literature require training both the backbone and the prototypical head.

By integrating our architecture with RAD-DINO, we achieved an AUROC of 86.1 on the RSNA dataset, when RAD-DINO reached 88.4 using linear probing. While there is a slight performance gap, this result highlights the potential of our architecture to transfer prototypical models to more specialized datasets. This is especially significant as more VFMs are being developed in domains where interpretability is crucial, such as the clinical field.

\subsection{Interpretability evaluation}
\begin{figure}[t] 
    \centering
    \includegraphics[width=.85 \columnwidth]{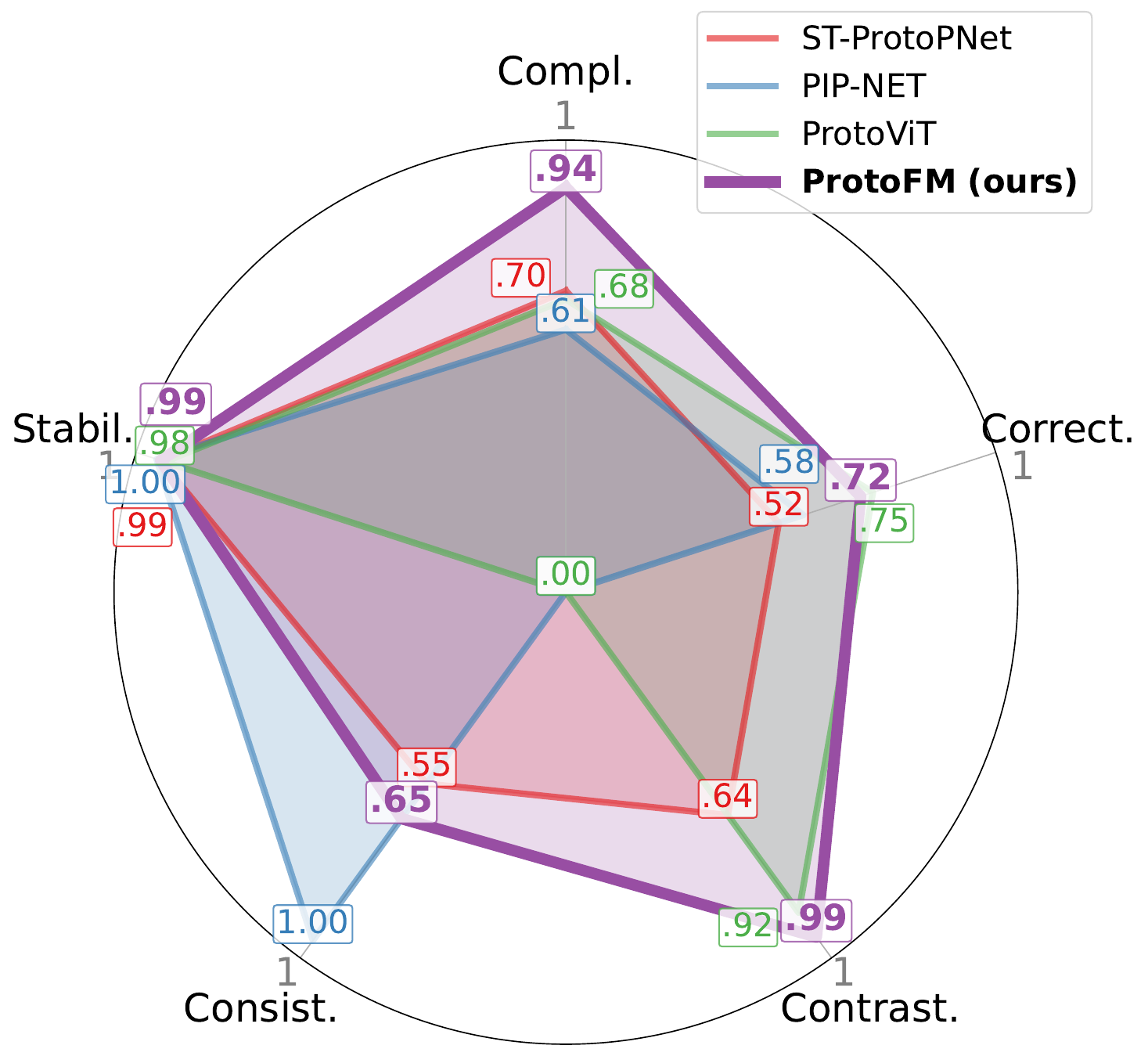}
    \caption{Radar plot summarizing model performance both in terms of Accuracy (Acc.) as well as explainability quality with the following metrics Global Size (Glob. Size), and Local Size (Loc. Size), Completeness (Compl.), Correctness (Correct.), and Contrastivity (Contrast.), Consistency (Consist.), and Stability (Stabil.).}
    \label{fig:radar_plot}
    \vspace{-5pt}
\end{figure}

We performed an extensive evaluation of the explanations provided by our model on the metrics described in \cref{sec:interp_metrics}. The model is compared with PIP-Net, ST-ProtoPNet and ProtoVit. These models were selected because ST-ProtoPnet and ProtoVit achieved the highest accuracy among prototypical models in the literature, while PIP-Net offered the most compact explanations in terms of both local and global size. 

The metrics presented in this section aim to cover the desiderata presented in~\cref{tab:interpretability_metrics}. Metrics which cover the first five desiderata are summarized in the radar plot shown in~\cref{fig:radar_plot}. The local and global sizes, which measure the compactness of the explanations, are reported in~\cref{tab:accuracy} and all metrics are reported individually in~\cref{sec:table_interp_metrics}.

The evaluation performed as part of our work demonstrates the multi-faceted nature of interpretability. As recognized in the literature, explanations produced by different models can be quite deceptive, and it is, therefore, important to perform a strong quantitative evaluation to ensure the explanations reflect at best the model's behavior. Our analysis highlighted different failure modes for two recently published models. 

Regarding PIP-Net, our analysis revealed that its prototypes often fail to highlight the discriminative features the model used to classify an image. Notably, PIP-Net scored zero on the target sensitivity metric, introduced by \cite{hesse2023funnybirds}. This metric quantifies the model’s reliance on known class-discriminative features in a synthetic dataset. This finding underscores the limitations of interpretability assessments based solely on human analysis. Although PIP-Net produces highly consistent prototypes, as evidenced by both the qualitative evaluation and the consistency metric in~\cref{fig:radar_plot}, our analysis also identifies specific failure modes.

Regarding ProtoVit, our analysis indicates that the model performs well across a range of benchmark metrics. However, we observed that the prototypes exhibit considerable inconsistency in the input space, with individual prototypes often being assigned to different bird parts. This inconsistency significantly lowers the model’s performance on the consistency metric. Additionally, the architecture lacks a sparsity constraint, resulting in a high local size. For instance, on the FunnyBirds dataset, an average of 504 prototypes have non-zero importance.  Low sparsity is a common issue in prototypical models.  Most score sheets include only three to four prototypes, yet the cumulative importance of these prototypes often represents only a small fraction of the final prediction. To partially address this issue, we introduce the Score Explained by Composition (SEC) metric, which we advocate for inclusion in all prototypical model score sheets. The SEC metric quantifies the extent to which the final prediction is explained by the prototypes presented in the score sheet, thereby improving transparency.

The proposed architecture, ProtoFM, achieves a mean explainability score (mX) of 0.92 based on the metrics defined in the FunnyBirds framework. This score is not limited to prototypical models, enabling comparisons across various interpretability approaches, including post-hoc methods. Our architecture outperforms all prototypical models evaluated in this paper, as well all  24 combinations of models and interpretability methods evaluated by~\cite{hesse2023funnybirds}. Regarding the metrics specific to prototypical model, our architecture scored 0.99 on the stabilit metrics  but attained a consistency score of 0.65, ranking higher than ST-ProtoPNet and ProtoViT but lower than PIP-Net which we observed produced very consistent prototypes. A user-study on CUB was performed to better understand the consistency of the prototypes generated by ProtoFM and all results are discussed in~\cref{app:User-Studies}.

One design choice of our architecture is to not push prototypes to a specific patch of the training set. This choice is supported by the neuroscience literature which  propose two models for concept representation: (i) the
exemplar model, where concepts are represented by multiple exemplars, and (ii) the prototype model,
where concepts are abstracted from specific exemplars~\cite{zeithamova2019brain}. Forcing prototypes to match specific
patches fails to align with either the exemplar or prototype model of concept representation in human
cognition. In this work, we take the first approach and represent concepts through exemplars as shown in~\cref{fig:sim_proto}. Instead of pushing the prototypes to a specific patch of the training set, we enforce all patches to be strongly assigned to a prototype through the alignment loss $L_{al}$. We further discuss the effects of the different losses in the next section.

\subsection{Ablation studies}
\begin{table}[t]
\caption{Ablation results on the Funny Birds dataset. Term from the loss objective are individually removed. Acc. stands for Accuracy, Conc. for consistency, $mX$ for the mean interpretability score and Loc. size for the local size}
\label{tab:ablation_funny}
\begin{center}
\begin{small}
\begin{tabularx}{0.99 \columnwidth}{cccc|cccY}
\toprule
$\mathcal{L}_{as}$ & $\mathcal{L}_{al}$& $\mathcal{L}_{NCE}$ & $\mathcal{L}_{sp}$ & Acc &Conc. & mX & Loc. \newline size \\ \midrule
         &      \checkmark     &  \checkmark      &   \checkmark        & \textbf{96.2 }       &    0.41         & 0.90   &     9       \\
     \checkmark     &           &    \checkmark     &  \checkmark          & 96.0        &    0.37         &   0.90 &       8     \\
   \checkmark       &         \checkmark   &         &   \checkmark         & 95.2        &     0.61       &  0.92  &        7    \\
      \checkmark    &     \checkmark       &     \checkmark    &           & 94.6        &     0.58        &   0.92 &        9    \\ \rowcolor{gray!15}
      \checkmark    &      \checkmark      &   \checkmark      &     \checkmark       & 95.8        &     \textbf{0.65 }       & \textbf{0.92 }  &         \textbf{6}   \\ \bottomrule
\end{tabularx}
\end{small}
\vspace{-5pt}
\end{center}

\end{table}

An ablation study to understand the effects of the different terms in the objective function was performed and the results are presented in~\cref{tab:ablation_funny}. This study was conducted on FunnyBirds by individually removing the different terms from the loss function and evaluating the model both in terms of classification performance and interpretability metrics. 

We find that the assignment loss $\mathcal{L}_{as}$ and the alignment loss $\mathcal{L}_{al}$ play a crucial role in enhancing prototype consistency. The significance of 
$\mathcal{L}_{al}$ , which leverages the similarity of features extracted by the backbone, further supports the decision to keep the backbone frozen—not only for efficiency but also for its contribution to consistency. Additionally, we observe that the contrastive loss $\mathcal{L}_{NCE}$ positively influences prototype consistency, while the sparsity loss $\mathcal{L}_{sp}$ contributes to reducing the local size of the explanation and hence promotes a better interpretability of the model for a given accuracy.

\section{Conclusion}
This work aimed to demonstrate that the proposed architecture ProtoFM effectively adapts visual foundation models into self-explainable classifiers. Through extensive evaluation, we showed that our approach not only produces models with competitive classification accuracy but also surpasses other prototypical models in the quality of explanations provided. While we believe this interpretability framework enhances understanding of the model’s decision-making process, a natural next step would be incorporating textual descriptions to further clarify the concepts utilized by the model.

% Additionally, exploring contrastive explanations could provide further insights into the model’s reasoning.

\section*{Impact Statement}

This paper presents work whose goal is to advance the field of 
Machine Learning (ML). There are many potential societal consequences of our work; however, we believe that interpretability, which is the focus of this work, could help to partially mitigate some of the risks of using ML-based models in critical applications.

\section*{Acknowledgments}
HT, MB and CL acknowledge financial support from the \textit{Fondation Carlos et Elsie De Reuter} and \textit{Fondation Ceres}. MB acknowledges support from Nicolas Pictet. GM acknowledges support from
MOE Tier 1 grant no. 22-4900-A0001-0: "Discipline-Informed Neural Networks for Interpretable
Time-Series Discovery". The computations were performed at University of Geneva using Baobab HPC service.

\bibliography{main}
\bibliographystyle{icml2025}

%%%%%%%%%%%%%%%%%%%%%%%%%%%%%%%%%%%%%%%%%%%%%%%%%%%%%%%%%%%%%%%%%%%%%%%%%%%%%%%
%%%%%%%%%%%%%%%%%%%%%%%%%%%%%%%%%%%%%%%%%%%%%%%%%%%%%%%%%%%%%%%%%%%%%%%%%%%%%%%
% APPENDIX
%%%%%%%%%%%%%%%%%%%%%%%%%%%%%%%%%%%%%%%%%%%%%%%%%%%%%%%%%%%%%%%%%%%%%%%%%%%%%%%
%%%%%%%%%%%%%%%%%%%%%%%%%%%%%%%%%%%%%%%%%%%%%%%%%%%%%%%%%%%%%%%%%%%%%%%%%%%%%%%
\newpage
\appendix
\onecolumn
\section{Additional materials}
\label{sec:additional_materials}
Additional materials are available on \href{https://zenodo.org/records/14778494?token=eyJhbGciOiJIUzUxMiJ9.eyJpZCI6ImYxNzVjNjQ0LTUwM2QtNGMzYS04M2IxLTQzODYxMjFlZjBmNiIsImRhdGEiOnt9LCJyYW5kb20iOiIxNDc2NjQzM2NmMzk2MzU5MGI4NDU5YzdhYTRiMmMxYiJ9.Z8cH0SyRhgruVVMUxRsQtKW8H4PpuKqssEKobSJ6Rk29nfPHDcCPwwkAd9nZNW15m9NXM8MBo-0T78AhIcMLwQ}{Zenodo}.

\section{Experimental Setup}
\label{sec:experimental_setup}
\subsection{General dataset}
The proposed architecture was implemented in PyTorch with all experiment trained on a on a single NVIDIA GeForce
RTX 3090, 12 cores and 64 GB of memory.  Models were trained for 120 epochs for PETS, CARS and CUB. For FunnyBirds, the model was trained for 50 epochs. All models were trained with an AdamW optimiser and the learning linearly increased during the first five epochs.
After the warm-up, the learning rate was progressively decreased following a cosine-decay schedule. Model's hyper-parameters are listed in~\cref{table:hyper_general}.

\begin{table}[h]
\caption{Parameter Settings for ProtoFM on the general dataset}
\label{table:hyper_general}
\centering
\begin{tabular}{cc}
\toprule
Parameter                         & Value \\ \midrule
Batch Size                        & 128   \\
Base Learning Rate                & 0.01  \\
Weight Decay                      & 0.01  \\
Number of Prototype $N$           & 300   \\
Image Size                        & 224   \\
teacher momentum                  & 0.995 \\
lr multiplier classification head & 10    \\
$\lambda_1 $                      & 2     \\
$\lambda_2 $                          & 5     \\
$\lambda_3 $                          & 1     \\
$\lambda_4 $                          & 0.1   \\
$\lambda_5 $                           & 2     \\
$\alpha$                             & 0.1   \\
$\gamma$                             & 0.1   \\ \bottomrule
\end{tabular}
\end{table}

\subsection{RSNA dataset}
We pre-process the RSNA dataset following the instructions provided by~\cite{perez2025exploring}. Images are initially resized using B-spline interpolation such that the shorter side is equal to 518. Min-max scaling is used to rescale the pixels in the range [0,255] and images are then saved in PNG. For training the image processor attached to RAD-DINO on hugging face is used. All hyper-parameters are kept the same, except the image size set to 518, $\lambda_5$ equals to 5 and the model being trained for 25 epochs.

\section{Alignment loss}
\label{sec:details_alignment_loss}
For the alignment loss, we followed the method by~\cite{kim2024eagle}, which updates the shift $b_{intra}$ and $b_{inter}$ for comparison with respective intra-sample features correlation and inter-sample correlation through the training as follows: \begin{align}
b_{\text{intra}} &= \left| \frac{1}{I} \sum_{i=1}^{I} \mathbf{\hat{F}}_{I}- \frac{1}{I} \sum_{i=1}^{I} \mathcal{\hat{A}}_{I} - k_{\text{shift}} \right| \\
b_{\text{inter}} &= \left( \frac{1}I \sum_{i=1}^{I}  \check{\mathbf{F}}_{I} + \frac{1}{I} \sum_{i=1}^{I}  \mathcal{\hat{A}}_{I} - k_{\text{shift}} \right) \times v_{\text{shift}},
\end{align}
$k_{\text{shift}}$ and $v_{\text{shift}}$ are set to 0.1 and 3.

This loss was used rather than the one from SmooSeg~\cite{smooseg} which collapses in our case given it can be equal to zero if all patches are assigned to the same prototype. Indeed it mutiplies the correlation matrix from the backbone with one minus the correlation of the prototypes assignment. Therefore in the case where all patches are assigned to the same prototype, the loss is equal to zero. 

\section{Additional details on interpretability metrics}
\label{sec:details_interp_metric_method}
We first adapt the part importance function $PI$ which aims to determine the importance of each part and is used to compute the metrics from the FunnyBirds framework. We align the redistribution of explanation on the input space such that our explanation follows the additive properties~\cite{scott2017unified}. Given the prototypes assignment matrix $\mathcal{A}$, weight from the classification head $W$, and the vector $h$ which represent the activation of the prototype, $PI$ for a given class $d$ at location $i$ is equal to: \begin{equation}
    PI[i,d] =  \sum_n \mathcal{A}_{i,n} \cdot W_{d,n} \cdot h_n \cdot \frac{1}{\sum_i \mathcal{A}_{i,n}}
\end{equation} 

Regarding the stability and consistency metrics, they are both based for each image on the vector $o_\textbf{n}$. This vector is a binary vector indicating whether prototype $p_n$ is related to category $u\in U$. There are five different parts categories for the FunnyBirds dataset: beak, eye, foot, tail and wing. For each category, we set the entry of the vector $o_\textbf{p}$ to one if an entry of the attribution matrix $\mathcal{A}_n$ weighted by the importance of the corresponding prototype $r_{d,n}$ is larger than 0.1 within the  binary segmentation mask corresponding to the given category $N_u$, where $N_u$ is equal to one if  the category is present at this location and zero otherwise:
\begin{equation}
    o_{\textbf{p}_n}^u =\max \left \{r_{d,n} \left( \mathcal{A}_n \circ N_u \right) \right \} >0.1
\end{equation}
The consistency and stability scores are then evaluated using the same equation as~\cite{huang2023evaluation} with our modified vector $o_\textbf{p}$. However as the initial paper considers prototypical models where prototypes only belong to one class, we repeat the operations across all classes.  Only the prototype that appears in the prediction for the considered class are included and  the result is averaged across all classes. 

\section{Interpretability metrics}
\label{sec:table_interp_metrics}
We present all metrics used to evaluate the model's interpretability along the metrics they are related to in~\cref{tab:interp_metrics_all}. 

\begin{table}[H]
\caption{Interpretability metrics for ProtoFM and three SOTA models. A stand for accuracy, with the next 6 columns referring to the metrics from~\cite{hesse2023funnybirds}, mX stand for the mean explainability score, while Con stands for consistency and Stab for stability as defined in the method of this paper. L.Size refers to the local size.}
\label{tab:interp_metrics_all}
\begin{center}
\begin{small}
\begin{tabular}{llllllllllllllll}
\toprule
            & A    & CSDC & PC   & DC   & D    & SD   & TS   & BI   & Com      & Cor  & Con  & mX   & Con  & Stab & L. Size \\ \midrule
PIP-Net      & 0.99 & 0.45 & 0.22 & 0.21 & 0.93 & 0.58 & 0.00 & 1.00 & 0.61 & 0.58 & 0.00 & 0.64 & 1.00 & 1.00 & 2   \\
ST-ProtoPNet & 1.00 & 0.78 & 0.69 & 0.67 & 0.69 & 0.52 & 0.64 & 1.00 & 0.70 & 0.52 & 0.64 & 0.77 & 0.55 & 0.99 & 20   \\
ProtoViT    & 0.94 & 0.97 & 0.93 & 0.96 & 0.40 & 0.75 & 0.92 & 1.00 & 0.68 & 0.75 & 0.92 & 0.86 & 0.00 & 0.98 & 504  \\
 \rowcolor{gray!15} ProtoFM     & 0.96 & 0.95 & 0.99 & 0.94 & 0.92 & 0.72 & 0.99 & 0.99 & 0.94 & 0.72 & 0.99 & 0.92 & 0.65 & 0.99 & 6    \\ \bottomrule
\end{tabular}
\end{small}
\end{center}
\end{table}

\section{Additional results}
\label{sec:additional_results}
The proposed architecture was also evaluated on PETS to compare its performance to the non-explainable baseline, DINO-B/14 as well as PIP-NET. Results are presented in~\cref{tab:results_pet}.

\begin{table}[h]
\caption{Classification accuracy along local and global size on PETS dataset}
\label{tab:results_pet}
\begin{center}
\begin{small}
\begin{tabular}{lccc}
\toprule
               & Accuracy {[}\%{]} & Local & Global \\ \midrule
DINO-B/14      & 96.2              &       &        \\
PIP-Net        & 92                & 2     & 172    \\
ProtoFM (ours) & 95.1              & 4     & 43     \\ \bottomrule
\end{tabular}
\end{small}
\end{center}
\end{table}

\subsection{Qualitative evaluation of prototypes consistency}
\label{app:User-Studies}

\begin{figure}[t!]
    \centering
    \begin{subfigure}[t]{0.45\textwidth}
        \centering
        \includegraphics[width=0.8\linewidth]{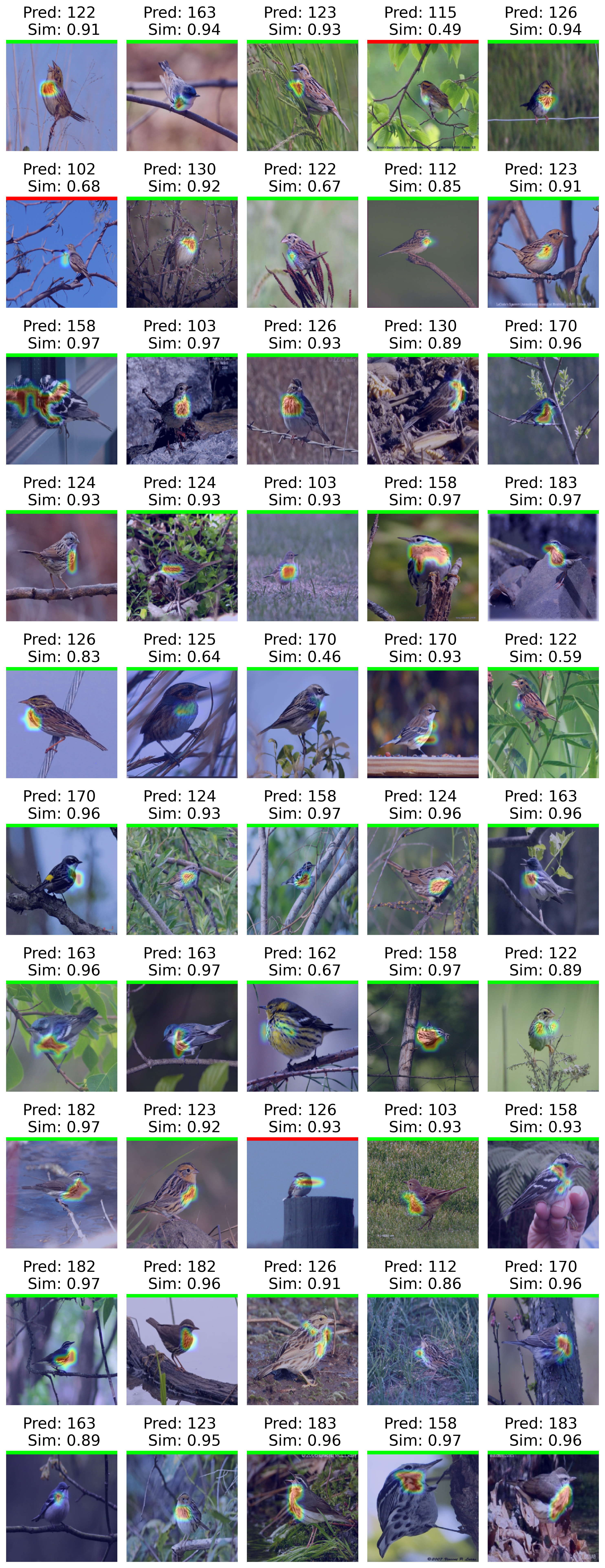}
        \caption{Prototype $26$: could be associated to "breast" part (100\% consistent).}
    \end{subfigure}%
    ~ 
    \begin{subfigure}[t]{0.45\textwidth}
        \centering
        \includegraphics[width=0.8\linewidth]{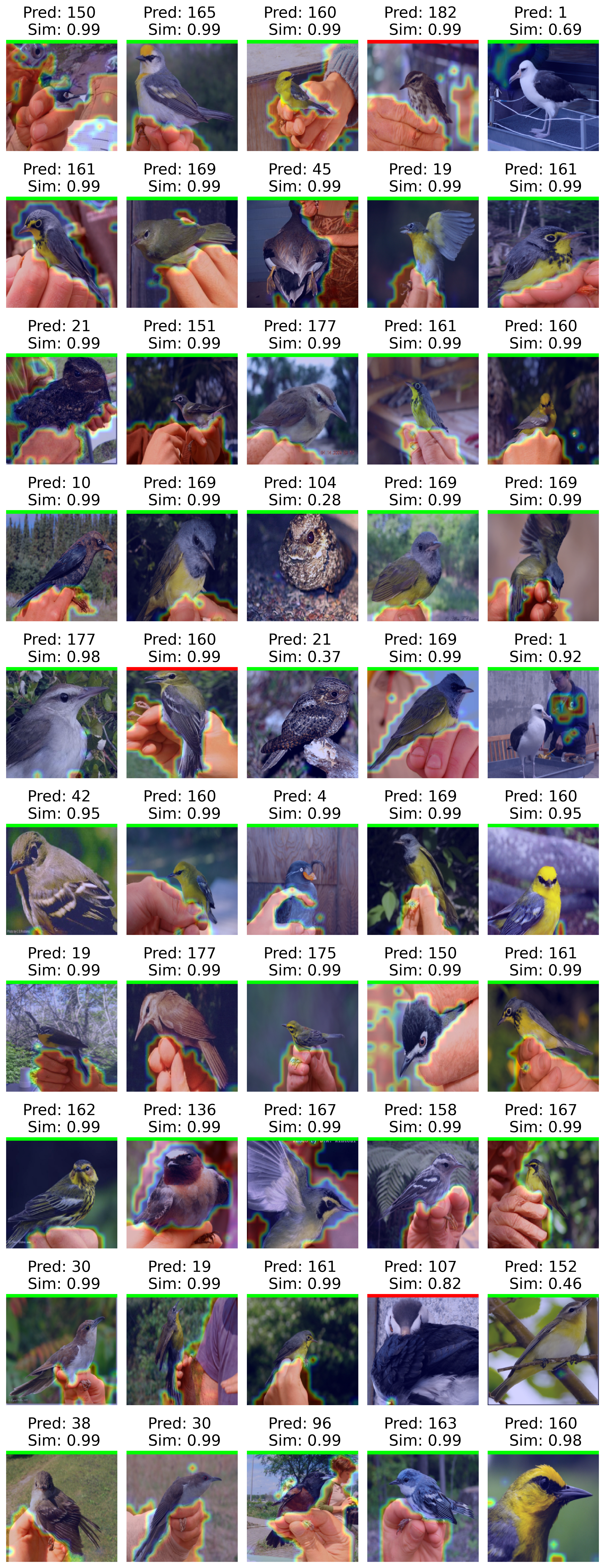}
        \caption{Prototype $108$: could be associated to "human" (82\% consistent).}
    \end{subfigure}
    \caption{Examples of random selection for each prototype of 50 samples where this prototype plays a role toward the model's prediction.}
    \label{fig:user_study}
\end{figure}

In addition to the quantitative metrics used to assess the quality of the explanations provided by the designed architecture, one user study was carried out to better understand the consistency of the prototypes with respect to concepts humans would associate together. The user study relies on a random selection for each prototype of 50 samples where this prototype plays a role toward the model's prediction.  The samples used for the user study can be found in Supplementary Materials~\cref{sec:additional_materials}.

A percentage of consistent sample (the most prevalent category) was calculated for each prototype (global size is 77). On average, the prototypes are 86\% consistent with one category (see analysis per prototype in Supplementary Materials (see~\cref{sec:additional_materials}), with 17 prototypes being 100\% consistent. Two examples are displayed in \cref{fig:user_study}. The model provides consistent explanations for fine-grained bird parts (such as "breast" or "beak") but is also able to produce more coarse-grained prototypes such as "human". One prototype (\# 81) identified a potential unwanted bias in the dataset, which are the textual parts present in the image (such as the signature).

This study focuses on a qualitative assessment only of the consistency of the prototypes, without considering the effects of displaying the explanation in the context of justifying a prediction for a specific class. As a result, protocols like HIVE~\cite{kim2022hive} are not applicable. Assessing the practical impact of the proposed explanation is beyond the scope of this paper. Evaluating its real-world implications, such as in clinical applications, would necessitate a complex user study incorporating protocols like HIVE alongside additional metrics that account for various factors influenced by actual clinical workflows.

\section{Additional visualizations}
\label{sec:additional_visualizations}
We present five score sheets per dataset in this section. Additional visualizations are available in the Additional materials.

\begin{figure}
    \centering
    \includegraphics[width=0.9\linewidth]{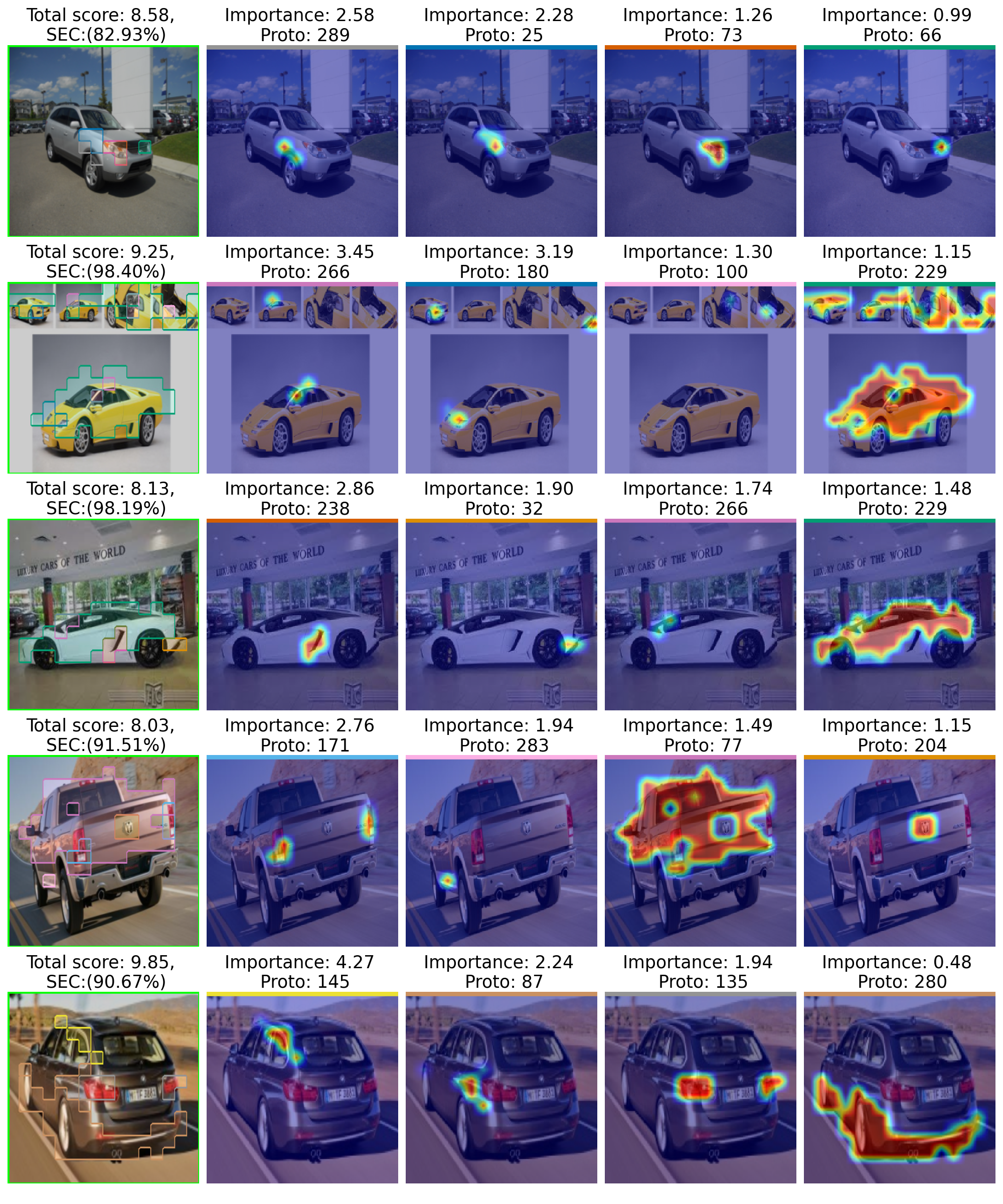}
    \caption{Score sheet for predictions on five random samples of the CARS dataset. Each row shows a
prediction on a different sample. The first column indicates the position of the top four prototypes.
Each subsequent column shows a prototype along with its importance towards the predicted class. The total score for the predicted class and the SEC metric are presented above the first column.}
    \label{fig:score_sheet_car_add}
\end{figure}

\begin{figure}
    \centering
    \includegraphics[width=0.9\linewidth]{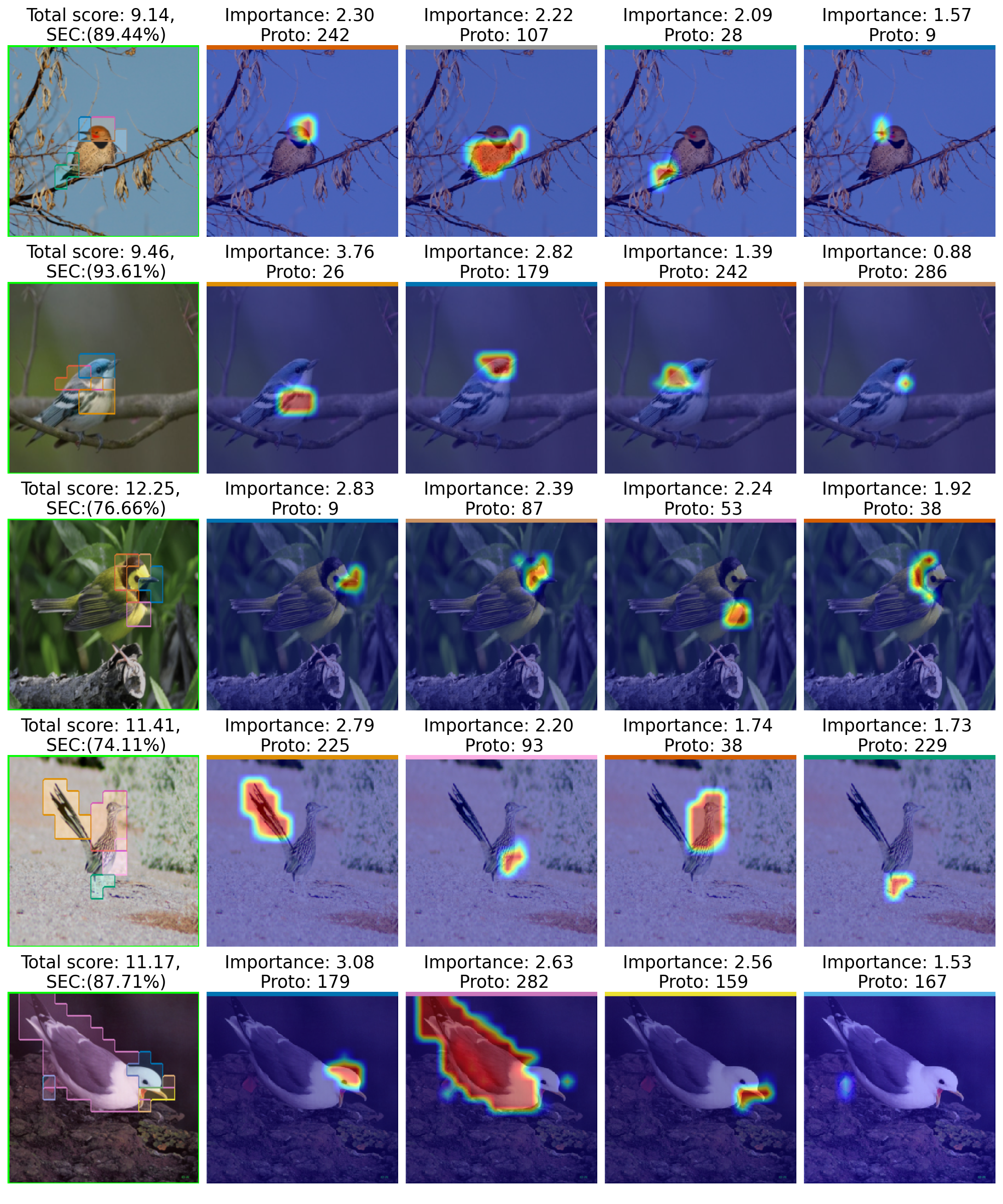}
    \caption{Score sheet for predictions on five random samples of the CUB dataset. Each row shows a
prediction on a different sample. The first column indicates the position of the top four prototypes.
Each subsequent column shows a prototype along with its importance towards the predicted class. The total score for the predicted class and the SEC metric are presented above the first column.}
    \label{fig:score_sheet_cub_add}
\end{figure}

\begin{figure}
    \centering
    \includegraphics[width=0.9\linewidth]{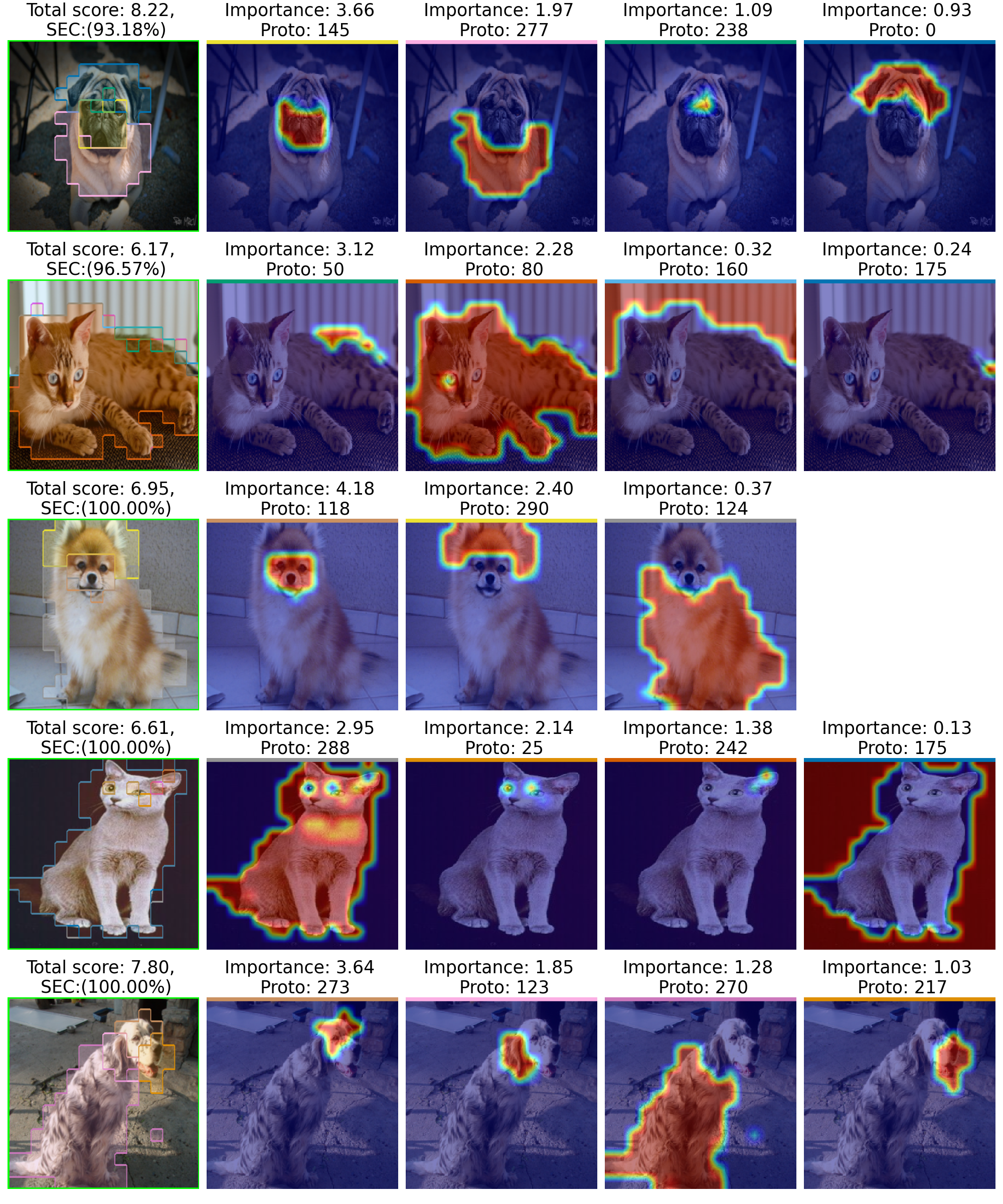}
    \caption{Score sheet for predictions on five random samples of the PETS dataset. Each row shows a
prediction on a different sample. The first column indicates the position of the top four prototypes.
Each subsequent column shows a prototype along with its importance towards the predicted class. The total score for the predicted class and the SEC metric are presented above the first column.}
    \label{fig:score_sheet_pet_add}
\end{figure}

\begin{figure}
    \centering
    \includegraphics[width=0.9\linewidth]{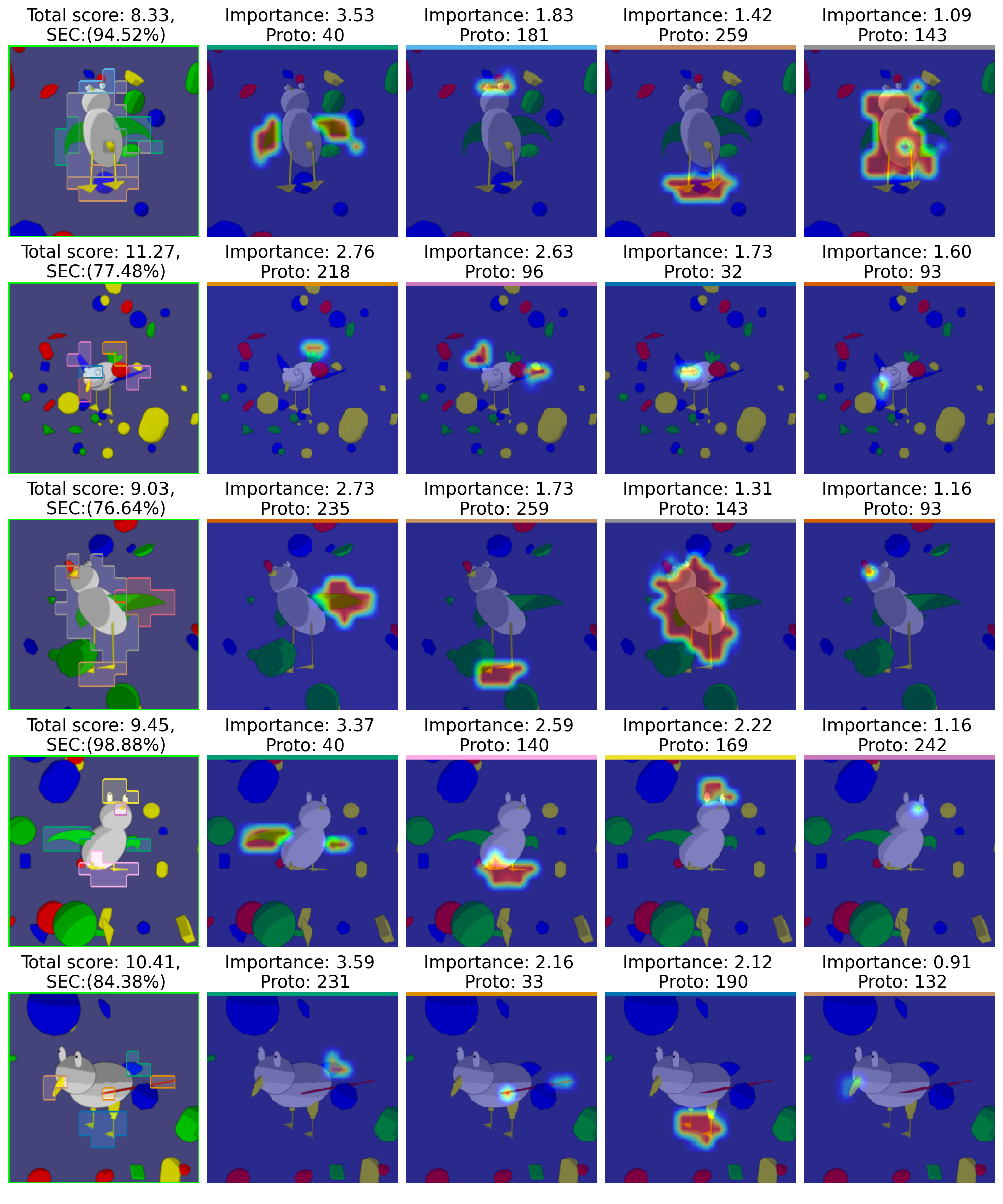}
    \caption{Score sheet for predictions on five random samples of the FunnyBirds dataset. Each row shows a
prediction on a different sample. The first column indicates the position of the top four prototypes.
Each subsequent column shows a prototype along with its importance towards the predicted class. The total score for the predicted class and the SEC metric are presented above the first column.}
    \label{fig:score_sheet_funnybirds_add}
\end{figure}

\begin{figure}
    \centering
    \includegraphics[width=0.9\linewidth]{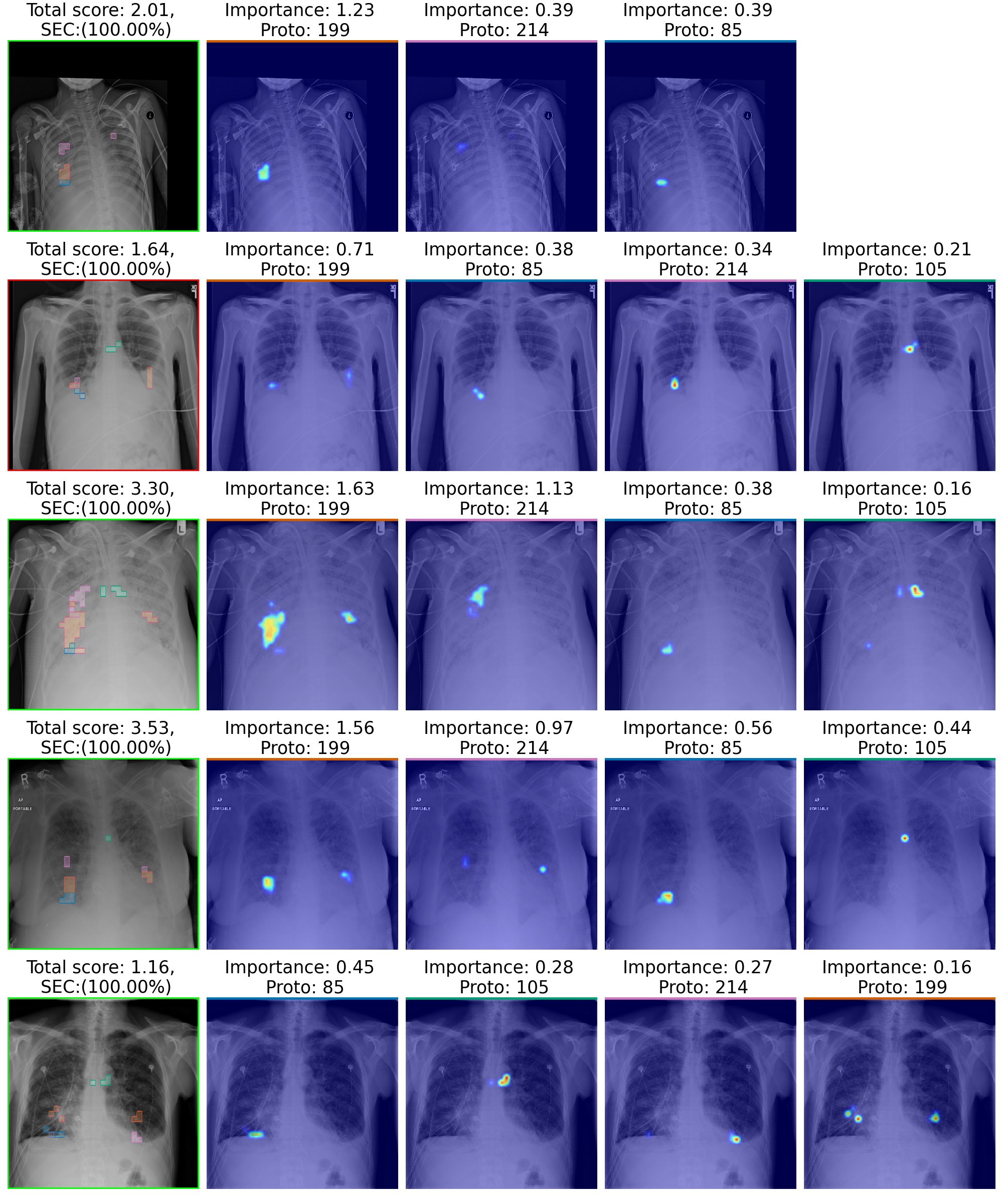}
    \caption{Score sheet for predictions on five random samples of the RSNA dataset. Each row shows a
prediction on a different sample. The first column indicates the position of the top four prototypes.
Each subsequent column shows a prototype along with its importance towards the predicted class. The total score for the predicted class and the SEC metric are presented above the first column.}
    \label{fig:score_sheet_rsna_add}
\end{figure}
%%%%%%%%%%%%%%%%%%%%%%%%%%%%%%%%%%%%%%%%%%%%%%%%%%%%%%%%%%%%%%%%%%%%%%%%%%%%%%%
%%%%%%%%%%%%%%%%%%%%%%%%%%%%%%%%%%%%%%%%%%%%%%%%%%%%%%%%%%%%%%%%%%%%%%%%%%%%%%%

\end{document}